\def\isarxiv{1} %%% for icml submission version, we comment this line

\ifdefined\isarxiv
\documentclass[11pt]{article}

\usepackage[numbers]{natbib}

\else
\documentclass{article}
\usepackage{neurips_2022}
\fi

\usepackage{amsmath}
\usepackage{amsthm}
\usepackage{amssymb}
\usepackage{algorithm}
\usepackage{subfig}
\usepackage{algpseudocode}
\usepackage{graphicx}
\usepackage{grffile}
\usepackage{wrapfig,epsfig}
\usepackage{url}
\usepackage{xcolor}
\usepackage{epstopdf}

\usepackage{bbm}
\usepackage{dsfont}

\allowdisplaybreaks

\ifdefined\isarxiv

\usepackage{tikz}
\usepackage{hyperref}  %%% arxiv don't allow this.
\hypersetup{colorlinks=true,citecolor=blue,linkcolor=blue} %%% Zhao : maybe we should comment this in submission.
\usetikzlibrary{arrows}
\usepackage[margin=1in]{geometry}

\else

\usepackage{microtype}
\usepackage{hyperref}
\definecolor{mydarkblue}{rgb}{0,0.08,0.45}
\hypersetup{colorlinks=true, citecolor=mydarkblue,linkcolor=mydarkblue}

\fi
\newtheorem{theorem}{Theorem}[section]
\newtheorem{lemma}[theorem]{Lemma}
\newtheorem{definition}[theorem]{Definition}

\newtheorem{fact}[theorem]{Fact}

\newtheorem{claim}[theorem]{Claim}

\newcommand{\wt}{\widetilde}

\newcommand{\N}{\mathcal{N}}
\newcommand{\R}{\mathbb{R}}

\DeclareMathOperator*{\E}{{\mathbb{E}}}
\DeclareMathOperator*{\var}{\mathrm{Var}}

\DeclareMathOperator{\poly}{poly}
\DeclareMathOperator{\asy}{asy}

\DeclareMathOperator{\diag}{diag}

\DeclareMathOperator{\dis}{dis}
\DeclareMathOperator{\cts}{cts}
\DeclareMathOperator{\Var}{Var}

\makeatletter
\newcommand*{\RN}[1]{\expandafter\@slowromancap\romannumeral #1@}
\makeatother

\usepackage{lineno}

\begin{document}

\ifdefined\isarxiv

\date{}

\title{An Over-parameterized Exponential Regression}
\author{
Yeqi Gao\thanks{\texttt{a916755226@gmail.com}. The University of Washington.}
\and
Sridhar Mahadevan\thanks{\texttt{smahadev@adobe.com}. Adobe Research.}
\and
Zhao Song\thanks{\texttt{zsong@adobe.com} Adobe Research.}
}

\else

\title{Intern Project} 
\maketitle 
\fi

\ifdefined\isarxiv
\begin{titlepage}
  \maketitle
  \begin{abstract}

Over the past few years, there has been a significant amount of research focused on studying the ReLU activation function, with the aim of achieving neural network convergence through over-parametrization. However, recent developments in the field of Large Language Models (LLMs) have sparked interest in the use of exponential activation functions, specifically in the attention mechanism.

Mathematically, we define the neural function $F: \R^{d \times m} \times  \mathbb{R}^d \rightarrow \mathbb{R}$ using an exponential activation function. Given a set of data points with labels $\{(x_1, y_1), (x_2, y_2), \dots, (x_n, y_n)\} \subset \mathbb{R}^d \times \mathbb{R}$ where $n$ denotes the number of the data. Here $F(W(t),x)$ can be expressed as $F(W(t),x) := \sum_{r=1}^m a_r \exp(\langle w_r, x \rangle)$, where $m$ represents the number of neurons, and $w_r(t)$ are weights at time $t$. It's standard in literature that $a_r$ are the fixed weights and it's never changed during the training. We initialize the weights $W(0) \in \mathbb{R}^{d \times m}$ with random Gaussian distributions, such that $w_r(0) \sim \mathcal{N}(0, I_d)$ and initialize $a_r$ from random sign distribution for each $r \in [m]$.

Using the gradient descent algorithm, we can find a weight $W(T)$ such that $\| F(W(T), X) - y \|_2 \leq \epsilon$ holds with probability $1-\delta$, where $\epsilon \in (0,0.1)$ and $m = \Omega(n^{2+o(1)}\log(n/\delta))$. To optimize the over-parametrization bound $m$, we employ several tight analysis techniques from previous studies [Song and Yang arXiv 2019, Munteanu, Omlor, Song and Woodruff ICML 2022].

  \end{abstract}
  \thispagestyle{empty}
\end{titlepage}

{\hypersetup{linkcolor=black}
\tableofcontents
}
\newpage

\else

\begin{abstract}

Over the past few years, there has been a significant amount of research focused on studying the ReLU activation function, with the aim of achieving neural network convergence through over-parametrization. However, recent developments in the field of Large Language Models (LLMs) have sparked interest in the use of exponential activation functions, specifically in the attention mechanism.

Mathematically, we define the neural function $F: \R^{d \times m} \times  \mathbb{R}^d \rightarrow \mathbb{R}$ using an exponential activation function. Given a set of data points with labels $\{(x_1, y_1), (x_2, y_2), \dots, (x_n, y_n)\} \subset \mathbb{R}^d \times \mathbb{R}$ where $n$ denotes the number of the data. Here $F(W(t),x)$ can be expressed as $F(W(t),x) := \sum_{r=1}^m a_r \exp(\langle w_r, x \rangle)$, where $m$ represents the number of neurons, and $w_r(t)$ are weights at time $t$. It's standard in literature that $a_r$ are the fixed weights and it's never changed during the training. We initialize the weights $W(0) \in \mathbb{R}^{d \times m}$ with random Gaussian distributions, such that $w_r(0) \sim \mathcal{N}(0, I_d)$ and initialize $a_r$ from random sign distribution for each $r \in [m]$.

Using the gradient descent algorithm, we can find a weight $W(T)$ such that $\| F(W(T), X) - y \|_2 \leq \epsilon$ holds with probability $1-\delta$, where $\epsilon \in (0,0.1)$ and $m = \Omega(n^{2+o(1)}\log(n/\delta))$. To optimize the over-parametrization bound $m$, we employ several tight analysis techniques from previous studies [Song and Yang arXiv 2019, Munteanu, Omlor, Song and Woodruff ICML 2022].

\end{abstract}

\fi

\section{Introduction}

Neural networks have proven to be effective in a range of different applications, such as image recognition \cite{ksh17, hzrs16} and speech recognition \cite{gmh13}. Overparametrization, the use of more parameters than necessary, is believed to be crucial to the success of deep learning \cite{als19a,als19b,dzps19,sy19,os20,syz21,bpsw21,mosw22,gqsw22}. Surprisingly, even when the data is improperly labeled and the target function is non-smooth and non-convex, over-parameterized neural networks trained with first-order methods can fit all training data, due to the modern architecture with ReLU activations. Furthermore, over-parameterized networks can improve generalization in practice, which contradicts traditional VC-dimension theory.

Large language models (LLMs) have proven to be more effective in processing natural language compared to smaller models and traditional algorithms. Examples of these models include Transformer \cite{vsp+17}, BERT \cite{dclt18}, GPT-3 \cite{bmr+20}, PaLM \cite{cnd+22}, and OPT \cite{zrg+22}.

The attention matrix is the key technical foundation of LLMs, as highlighted in previous research \cite{vsp+17, rns18, dclt18, rwc+19, bmr+20}. This square matrix has rows and columns that correspond to words or "tokens" in natural language, with entries representing the correlations between them. It is used to determine the importance of each token in a sequence when generating an output. In the attention mechanism, each input token is assigned a weight or score based on its relevance to the current output. These scores are calculated using a similarity function that compares the input and output states.

The exponential activation function \cite{dts15,sa19} is a commonly used activation function in neural networks. It maps the input to the output using the exponential function.
One of the main advantages of the exponential activation function is that it can produce positive outputs for any input, which can be useful in certain types of neural networks, such as those used for regression tasks. Additionally, the exponential activation function is continuously differentiable, which is important for backpropagation during training.

Another application of the exponential activation function is in the generation of natural language \cite{vsp+17,dclt18,bmr+20,cnd+22}. Specifically, it has been used in language models such as GPT-3 \cite{bmr+20} to generate text that closely mimics human writing. The exponential function can help to weight the importance of different words in a given context, leading to more accurate and coherent language generation. 

\cite{o23} present GPT-4, a multimodal model capable of producing text outputs from both image and text inputs. While GPT-4 falls short of human-level performance in some real-world scenarios, it achieves human-like results on various professional and academic benchmarks, including scoring in the top $10\%$ on a simulated bar exam. Based on the Transformer architecture, GPT-4 is pre-trained to predict the next token in a document. Post-training alignment improves its accuracy in factual content and adherence to desired behavior. \cite{o23} also developed infrastructure and optimization methods that scale predictably, allowing for accurate predictions of GPT-4's performance using models trained on just $1/1000$ th of its computational capacity.

In this work, we consider a natural question,
\begin{center}
{\it Is that possible to prove an over-parameterization bound for exponential activation in neural network learning?}
\end{center}
In this work, we provide a positive answer for this question. 
Assuming a set of data points $\{(x_i, y_i)\}_{i=1}^n \subset \mathbb{R}^d \times \mathbb{R}$, we use $\lambda$ as the minimum eigenvalue of the neural tangent kernel with respect to the exponential activation function, $n$ as the number of points and $m$ as the number of neurons. Moreover, we use $F(t)$ to represent our two-layer neural network with the exponential activation function at the $t$-th step.

Based on our analysis of the perturbation of the weights, we can derive a bound on the prediction loss $\|y-F(t)\|_2^2$. By choosing $m$ to be large enough, specifically $\Omega( \lambda^{-2} \log(n/\delta) n^{2+o(1)} )$, and setting the learning rate $\eta= \Theta( \lambda / ( m n^{2+o(1)} ) )$, we can ensure the convergence of the neural network with exponential activation.

\subsection{Our Results}

Our primary result is presented below.
\begin{theorem}[Main result, formal version of Theorem~\ref{thm:formal}]\label{thm:informal}
Let $\delta \in (0,0.1)$ denote the failure probability. Let $\epsilon \in (0,0.1)$ denote the accuracy.
If the following conditions hold
\begin{itemize}
    \item Let  $\lambda > 0$ denote the minimum eigenvalue of neural tangent  kernel with respect to exponential activation.
    \item Let $m = \Omega( \lambda^{-2}  \log(n/\delta)n^{2+o(1)} )$ represent the number of neurons.
    \item Let $w_r$ be random Gaussian weights and $a_r$ be the random $\{-1,+1\}$ weights.
    \item Let $\eta= \Theta( \lambda / ( m n^{2+o(1)}  ) )$ denote the learning rate of gradient descent algorithm
    \item Let $T =\Omega( \lambda^{-2} n^{2+o(1)}  \cdot \log(n/\epsilon) )$ denote the number of iterations of gradient descent algorithm 
\end{itemize}
Then, we have after running algorithm with $T$ iterations. And with probability at least $1-\delta$, we obtain a $w(T)$ such that
\begin{align*}
    \| F(T) - y \|_2^2 \leq \epsilon.
\end{align*}
\end{theorem}

In order to demonstrate the convergence, we begin by selecting a sufficiently large value of $m$ to regulate changes in weights $w$ and gradients over the training. We then assume that $w$ is contained within a small ball, allowing us to complete the proof of convergence.

\section{Related Work}

\subsection{Training over-parameterized neural network}
\paragraph{Convergence}

\cite{dzps19} demonstrate that for a shallow neural network with ReLU activation consisting of $m$ hidden nodes and trained on $n$ data points, as long as $m$ is sufficiently large and no two inputs are parallel, randomly initialized gradient descent will converge to an optimal global solution at a linear rate of convergence for the quadratic loss function. This is due to the fact that over-parametrization and random initialization work together to ensure that every weight vector remains close to its initial value throughout all iterations.

\cite{als19a} have proven that under two assumptions, simple algorithms such as stochastic gradient descent (SGD) can discover Global Minima on the training objective of Deep Neural Networks (DNNs) in Polynomial Time. The two assumptions are that the inputs are non-degenerate, and the network is over-parameterized, meaning the number of hidden neurons is sufficiently large and is polynomial in $L$, the number of DNN layers, and in $n$, the number of training samples. \cite{als19b} study recurrent neural networks (RNNs) used in natural language processing in \cite{als19b}. They demonstrate that with enough neurons, SGD can minimize the regression loss at a linear rate, showing RNNs can memorize data. \cite{als19b} also develop a perturbation theory for analyzing first-order approximation of multi-layer networks.

\paragraph{Over-parametrization bound, bound on $m$}
In deep learning theory, \cite{sy19} enhance the size of over-parametrization beyond three previous notable results \cite{ll18}, \cite{dzps18}, and \cite{adh+19}.

In neural network training, it is common to initialize all weights as independent Gaussian vectors. However, \cite{mosw22} observed that initializing the weights as independent pairs, where each pair consists of two identical Gaussian vectors, can improve the convergence analysis significantly.

\paragraph{Using data structure to speedup cost per iteration}

Preprocessing plays a critical role in the training of over-parameterized neural networks. \cite{syz21} demonstrates that the cost per iteration of training can be reduced by pre-processing the initial weights of the neural network or pre-processing the input data points. Specifically, pre-processing the initial weights can result in a cost of $\wt{O}(m^{1-\Theta(1/d)}nd)$ per iteration, while pre-processing the input data points can further reduce the cost to $\wt{O}(m^{4/5}nd)$ per iteration. \cite{als+22} also propose a new preprocessing method that employs a tree data structure to detect neuron firing during each iteration and  achieves $o(nmd)$ time per iteration and requires $O(nmd)$ time in preprocessing. By using $m^2$ cost only in the initialization phase, \cite{szz21,z22} reach the cost of  $m^{2-\Omega(1)}$ per iteration.

Some works focused on the faster second-order optimization algorithms. Although second-order algorithms have a remarkable convergence rate, their high computational cost per iteration renders them impractical. Recent work by \cite{zmg19, cgh+19} which focused on the second-order algorithms has mitigated this computational overhead, resulting in an $O(mn^2)$-time second-order algorithm for training two-layer over-parameterized neural networks. \cite{bpsw21} further accelerates the algorithm of \cite{cgh+19} to achieve an $\wt{O}(mn)$-time backpropagation algorithm for training mildly over-parameterized ReLU networks.

By utilizing data structures, certain methods can decrease the cost per iteration, resulting in faster performance. \cite{gqsw22} analyze the convergence guarantee of adversarial training on a two-layer neural network with shifted ReLU activation, finding that only $o(m)$ neurons are activated per input data per iteration which reached the training cost time cost of $o(mnd)$ per iteration.  \cite{hswz22} introduce a novel training approach for a standard neural network with $m = \poly(n)$ parameters and a batch of $n$ input data points in $\R^d$. By treating neural networks as a collection of binary search trees and making selective modifications to a subset of nodes at each iteration, with $\alpha \in (0.01,1)$ fixed, their method achieves a time complexity of $m^{1-\alpha}nd+n^3$ in the overparametrized regime.

\subsection{Attention Theory}

\paragraph{Fast computation and optimization}

In the field of optimazation, \cite{zkv+20} focused the role of adaptive methods in attention models and by \cite{szks21} focused on analyzing the dynamics of a single-head attention head to approximate the learning of a Seq2Seq architecture.
According to \cite{grs+23}, the attention mechanism is a versatile tool that can be used to execute complex, general-purpose programs even in shallow transformer models. \cite{zkv+20} for the role of adaptive methods in attention models and by \cite{szks21} for analyzing the dynamics of a single-head attention head to approximate the learning of a Seq2Seq architecture.

The computation of attention is a crucial aspect of training large language models. Given three matrices $Q, K, V \in [-B, B]^{n\times d}$ as input, the objective is to construct the matrix $\textsc{Att}(Q, K, V ) := \diag( A {\bf 1}_n)^{-1}AV \in \R^{n\times d}$, where $A =\exp(QK^{\top }/d)$ is the ‘attention matrix’. In \cite{as23}, the authors investigate whether faster algorithms are possible by implicitly utilizing the matrix $A$. They present two results that show a sharp transition occurs at $B=\Theta(\sqrt{\log n})$.
A recent study conducted by Zandieh, Han, Daliri, and Karbasi \cite{zhdk23} introduced the first algorithm with provable guarantees for attention approximation. Their algorithm employs techniques from locality sensitive hashing (LSH) \cite{ckns20}. \cite{zhdk23,as23} are mainly focusing on the static version of attention computation problem. \cite{bsz23} define and study the dynamic version of attention computation problem. \cite{bsz23} provides both algorithmic result and hardness result. \cite{rkr+23} introduce the Structured State Space (S4) sequence model, which utilizes a novel parametrization for the SSM. The study demonstrates that the S4 model can be computed with significantly greater efficiency than previous approaches, while still retaining their theoretical advantages. \cite{lsz23} study the regularized exponential regression problem, and provides an algorithm that runs in input sparsity time.

\paragraph{Expressivity for transformer}

 Expressivity has been studied by \cite{jadc20,egkz22} for self-attention blocks and by \cite{dgv+18,wcm22,hbsn20} for Transformers. 
Research has demonstrated that fine-tuning language models on a set of datasets expressed as instructions can enhance model performance and improve generalization to unfamiliar tasks. \cite{chl+22} investigate instruction fine-tuning with a specific emphasis on expanding the number of tasks, increasing the size of the model, and  fine-tuning on chain-of-thought data. \cite{egkz22} offers a thorough theoretical examination of the inductive biases associated with self-attention modules. The goal is to establish, with rigor, the types of functions and long-range dependencies that self-attention blocks are inclined to represent.

A recent study conducted by \cite{lag+22} delved into this matter by exploring learning automata, which are discrete dynamic systems that are well-suited for recurrent modeling and expressing algorithmic tasks.

The study conducted by \cite{onr+22} commences by introducing a simple weight construction that establishes the likeness between data transformations generated by  a single linear self-attention layer and  gradient-descent (GD) implemented on a regression loss. \cite{llr23} offer a detailed mechanistic explanation of how transformers learn "semantic structure," defined as the ability to capture the co-occurrence patterns of words.

\paragraph{In-context learning}
Many works focused on the in-context learning in recent years. 
While beneficial, the quadratic complexity of self-attention on the
input sequence length has limited its application to longer sequences, a topic being actively studied in the community. To
address this limitation, \cite{xzc+21} propose Nystromformer, a model
that exhibits favorable scalability as a function of sequence
length.
During testing, in-context learning takes place as the language model (LM) deduces a common underlying concept among the examples given in a prompt. In a study by \cite{xrlm21}, it was demonstrated that this phenomenon can occur even when there is a difference in the distribution of prompts and the pretraining data. The study was conducted in a scenario where the pretraining data had a combination of hidden Markov models (HMMs).

\cite{asa+22} explore the possibility that transformer-based in-context learners implicitly execute standard learning algorithms by encoding smaller models within their activations and updating these implicit models as new examples are introduced in the context.

In order to gain a better understanding of in-context learning, \cite{gtlv22} examine the well-defined problem of training a model to in-context learn a function class (such as linear functions). Their study investigates whether a model can be trained to in-context learn "most" functions from this class when given data derived from some functions in the class.

The key finding of \cite{egkz22} demonstrates that Transformer networks with bounded-norm have the ability to "create sparse variables." Specifically, a single self-attention head can represent a sparse function of the input sequence, and the sample complexity scales logarithmically with the context length. \cite{mss22} demonstrate that saturated transformers surpass the known limitations of hard-attention transformers. \cite{mss22} subsequently establish that saturated transformers, which utilize floating-point values, can be replicated through constant-depth threshold circuits, which restricts the class of formal languages they can identify.

 \cite{kgw+23} introduce a watermarking framework designed specifically for proprietary language models. This watermark can be embedded with minimal impact on text quality, and can be detected using an efficient open-source algorithm that does not require access to the language model API or parameters.

\paragraph{Other applications and theories of transformer}

Some analysis focused on the Transformer computing on hardware. One of the key principles missing in attention algorithms is their lack of IO awareness, i.e., not accounting for the reads and writes between different levels of GPU memory. In order to address the aforementioned issue, the authors of the paper referenced as \cite{dfe+22} have introduced a new attention algorithm called FlashAttention. This algorithm is designed to be precise, taking into account input-output (IO) operations, while also leveraging tiling to minimize the number of times data needs to be transferred between on-chip SRAM and the GPU's high bandwidth memory (HBM).

\cite{beg+22} examines the ability of neural networks to learn a $k$-sparse parity of n bits, a well-known discrete search problem that is statistically simple but computationally difficult. Through empirical investigations, the authors observe that various neural networks are capable of successfully learning sparse parities, and they note discontinuous phase transitions in the training curves.

\cite{vkb23} propose modifications to generative model learning algorithms that provide strong bounds on the probability of sampling protected content. These modifications are made in an efficient and black box manner.

\paragraph{Roadmap.}

Our techniques are outlined briefly in Section~\ref{sec:tech_overview}, while Section~\ref{sec:preli} covers our preliminary tools and notations. In Section~\ref{sec:problem_formulation}, we introduce the problem of interest and define a two-layer neural network with exponential activation functions.
In Section~\ref{sec:intialization_and_perturbation}, we demonstrate that when the width $m$ is sufficiently large, the continuous and discrete versions of the input data's Gram matrix are spectrally close to each other. Section~\ref{sec:convergence} establishes that an over-parameterized neural network achieves linear convergence of the training error to $0$. In Section~\ref{sec:induction_for_weight}, we simplify the problem by defining $D_{\cts},H(s)$ and providing a gradient bound through induction. Then, in Section~\ref{sec:induction_for_loss}, we establish a similar induction-based bound for the loss $\|y-F(t+1)\|_2^2$ at any time.

  %%% Section 1. Introduction

\section{Technique Overview}\label{sec:tech_overview}
This paper presents a proof demonstrating that a two-layer neural network employing the exponential activation function which can achieve a desired small loss value after sufficient iterations, given a large enough number of neurons $m$, an appropriate learning rate $\eta$, and the initialization method specified in Definition~\ref{def:duplicate_weights}.

By bounding the difference of the weights over the training and choosing a proper learning rate $\eta$, we bound the loss by induction. We will introduce how we bound the loss under the assumption for the small perturbation on the weights. And then we will introduce how we bound the  weights and gradients respectively.

\paragraph{Bounding the loss by induction}

To establish this result, we begin by bounding the summation of the differences between the weights at the current step and their initial values, assuming that $w$ is in a small range such that $\Delta w_r(t)\leq R\in (0,0.01)$ where $t$ denote the step here. 

To establish an upper bound on the prediction loss $\|y-F(t+1)\|_2^2$, we first assume that the weight $w$ is within a small range, namely $\Delta w_r(t)\leq R\in (0,0.01)$. We decompose the loss into four parts: 
\begin{itemize}
    \item The loss at the previous step  $\|y-F(t)\|_2^2$
    \item $C_1:=-2m \eta (F(t)-y)^\top H(t) (F(t)-y)$
    \item $C_2:=2 (F(t)-y)^\top H_{\asy} (t) (F(t)-y)$
    \item $C_3:=\| F(t+1) - F(t) \|_2^2$
\end{itemize}

Using terms that involve the parameters $m$, $\eta$, $n$, and $B$ multiplied by $\|y-F(t)\|_2^2$, we can compute the upper bounds $C_1$, $C_2$, and $C_3$ respectively. Then, by induction and choosing a proper value for $m$, we can bound $\|y-F(t)\|_2^2$ where the loss is $\|y-F(0)\|_2^2= \|y\|_2^2$ at initialization under the given assumption.

\paragraph{Bounding the weights by induction}
Finally, to complete the proof, we establish an upper bound for $\Delta w_r(t)=\|w_r(t)-w_r(0)\|_2$. To do this, we transform $\Delta w_r(t)$ into a form that is a multiple of $\exp(B+R)\sqrt{n}$ and the current loss $\|y-F(t)\|_2$. 

\paragraph{Bounding the gradients by induction}
 Based on the result above, we will continue our work on bounding the changing of the weights over the training. By appropriately choosing the learning rate $\eta$, we can ensure that $\Delta w_r(t)$ is small enough, namely $0.01$. We have the following definition 
\begin{align*}
    H(w)_{i,j} :=  \frac{1}{m} \langle x_i,x_j\rangle \sum_{r\in [m]} \exp(\langle w_r,x_i\rangle)\cdot \exp( \langle w_r,x_j\rangle)
\end{align*}
where $r\in [m]$ as the index the neurons, and $x_i$ denotes the data where $i\in [n]$ and $j\in [n]$.

Next, we will constrain the changes of $H$ under the assumption that $w$ is located within a small ball. In addition, we must bound the discrepancy between discrete and continuous functions. Drawing upon the conclusions reached regarding perturbations in weight $w$, we can guarantee the convergence of the over-parametrized neural network with an exponential activation function.

\section{Preliminary}\label{sec:preli}

In Section~\ref{sec:preli:notations}, we provide several basic notations and definitions. In Section~\ref{sec:preli:data_points}, we present our assumptions for data points. 
In Section~\ref{sec:preli:weights}, we outline our assumptions regarding weight initialization. Section~\ref{sec:preli:algera} provides some basic algebraic concepts used throughout the paper. Finally, Section~\ref{sec:preli:probtools} discusses various probability tools used in our work.

\subsection{Notations}\label{sec:preli:notations}

In our notation, $[n]$ represents the set $\{1,2,\cdots, n\}$. The $\exp$ activation function is denoted by $\phi(x) = \exp(x)$.

The $\ell_2$ norm of a vector $y \in \R^n$ is denoted by $\| y \|_2:= (\sum_{i=1}^n y_i^2)^{1/2}$ and represents the element-wise square root of the sum of squares of each entry in the vector.

The spectral norm of a matrix $B$ is denoted by $\| B \|$. We also define the Frobenius norm $\| B \|_F = ( \sum_{i} \sum_{j} B_{i,j}^2 )^{1/2}$ and the $\ell_1$ norm $\| B \|_1 = \sum_{i} \sum_{j} |B_{i,j}|$ of matrix $B$.

For any symmetric matrix $B\in \R^{k\times k}$, we define its eigenvalue decomposition as $U\Lambda U^\top$, where $\Lambda$ is a diagonal matrix. Let $\lambda_1,\cdots ,\lambda_k$ denote the entries on diagonal of $\lambda \in \R^{k\times k}$. We say $\lambda_i$ is the $i$-th eigenvalue. Usually, we write it as $\lambda_i(B)$ where $i\in [k]$.

We define 
\begin{align*}
    \lambda_{\min} (B) := \min_{i\in [k]}\{\lambda_1,\cdots,\lambda_k\}.
\end{align*}

We use ${\cal N}(\mu, \Sigma)$ to denote a $d$-dimensional Gaussian distribution with mean $\mu \in \R^d$ and covariance matrix $\Sigma \in \R^{d \times d}$.

\subsection{Data points}\label{sec:preli:data_points}

\begin{definition}\label{def:data}
We assume the data points satisfy 
\begin{itemize}
    \item $\| x_i \|_2 \leq 1$, for all $i \in [n]$
    \item $|y_i | \leq 1$, for all $i \in [n]$
\end{itemize}
\end{definition}

\subsection{Initialization Weights}\label{sec:preli:weights}
The following weights initialization are very standard in the literature, e.g., see \cite{dzps19,sy19,bpsw21,syz21}.
\begin{definition}\label{def:standard_weights}
We choose weights as follows
\begin{itemize}
    \item $\forall r\in [m]$, $a_r$ is randomly and uniformly sampled from the set $\{-1, +1\}$.
    \item We sample $w_r$ from ${\cal N}(0, \sigma^2 I)$ for each $r \in [m]$
\end{itemize}
\end{definition}

To improve the initialization bound $\| y - F(0) \|_2^2$, we will use an idea from \cite{mosw22}. 
The definition~\ref{def:standard_weights} and Definition~\ref{def:duplicate_weights} are the same up to constant factor. So for convenient of analysis, in most places we use Definition~\ref{def:standard_weights}. We only use Definition~\ref{def:duplicate_weights} for bounding the initialization $\| y - u (0) \|_2^2 = 0$.

\begin{definition}\label{def:duplicate_weights}
For each $r\in [m/2]$, we choose weights as follows
\begin{itemize}
    \item  We  sample $a_{2r-1}$ from $\{-1,+1\}$ uniformly at random. 
    \item We sample $w_{2r-1}$ from Gaussian distribution ${\cal N}(0, \sigma^2 I)$ .
    \item We choose $a_{2r} = -a_{2r-1}$.
    \item We choose $w_{2r-1} = w_{2r}$.
\end{itemize}
\end{definition}

\subsection{Basic Algebra}\label{sec:preli:algera}

\begin{fact}[Taylor series]\label{fac:exact_taylor}
We have
\begin{itemize}
    \item $\exp(x) = \sum_{i=0}^{\infty} \frac{1}{i!} x^i$
    \item $\cosh(x) = \sum_{i=0}^{\infty} \frac{1}{(2i)!} x^{2i} $
    \item $\sinh(x) = \sum_{i=0}^{\infty} \frac{1}{(2i+1)!} x^{2i+1} $
\end{itemize}

\end{fact}

\begin{fact}[Cauchy Schwarz]\label{fac:cauchy_schwarz}
For any two vectors $x,y \in \R^n$, we have 
\begin{align*}
\langle x, y \rangle \leq \|x \|_2 \cdot \| y \|_2.
\end{align*}
\end{fact}

\begin{fact}\label{fac:norm}
We have
\begin{itemize}
    \item $\| B  \| \leq \| B \|_F$  
    \item  $\forall B\in \R^{n \times n}$, $\| B \|_F \leq n \| B \|_{\infty} $
    \item  $\forall x \in  R^n$,  $x^\top B x \leq \| x \|_2^2 \cdot \| B \|$
    \item $\lambda_{\min}(A) \geq \lambda_{\min}(B) - \| A - B \| $  
\end{itemize}

\end{fact}

\begin{fact}\label{fac:taylor}
We have
\begin{itemize}
    \item For any $ |x| \leq 0.1$, then we have $| \exp(x) - 1 | \leq 2x$.
    \item For any $|x| \leq 0.1$, then we have $|\cosh(x) - 1| \leq x^2$
    \item For any $|x| \leq 0.1$, then we have $\exp(x) = 1+x + \Theta(1) x^2$.
    \item For any $|x| \leq 0.1$, then we have $(1-x)^{1/2} \leq 1-0.5 x$
    \item For any $x \in (0,0.1)$, we have $\sum_{i=0}^{\infty} x^i \leq \frac{1}{1-x}$
\end{itemize}
\end{fact}

\subsection{Probability Tools}\label{sec:preli:probtools}

We state the standard Bernstein inequality,
\begin{lemma}[Bernstein inequality \cite{b24}]\label{lem:bernstein}
If the following condition holds
\begin{itemize}
    \item $Z_1, \cdots, Z_n$ be independent zero-mean random variables
    \item $|Z_i| \leq M$ almost surely $\forall i\in [n]$
    \item Let $Z=\sum_{i=1}^n Z_i$.
    \item $\Var[Z] = \sum_{j=1}^n \E[Z_j^2]$.
\end{itemize}
 Then, for all positive $t$,
\begin{align*}
\Pr \left[ Z > t \right] \leq \exp \left( - \frac{ t^2/2 }{ \var[Z]  + M t /3 } \right).
\end{align*}
\end{lemma}

We state the standard Hoeffding inequality,
\begin{lemma}[Hoeffding inequality \cite{h63}]\label{lem:hoeffding}
If the following conditions hold
\begin{itemize}
    \item Let $Z_1, \cdots, Z_n$ denote $n$ independent variables
    \item $Z_i \in [\alpha_i,\beta_i]$, for all $i \in [n]$
    \item  Let $Z= \sum_{i=1}^n Z_i$.
\end{itemize}
 Then we have
\begin{align*}
\Pr[ | Z - \E[Z] | \geq t ] \leq 2\exp \left( - \frac{2t^2}{ \sum_{i\in [n]} (\beta_i - \alpha_i)^2 } \right).
\end{align*}
\end{lemma}

We state a standard tool from literature (see Lemma 1 on page 1325 of \cite{lm00}),
\begin{lemma}[Laurent and Massart \cite{lm00}]\label{lem:lm}
Suppose $X$ follows a chi-squared distribution with $k$ degrees of freedom denoted by $\mathcal{X}_k^2$. The mean of each variable is zero, and the variance is $\sigma^2$.
    Then,
    \begin{align*}
        \Pr[X - k\sigma^2 \geq (2\sqrt{kt} + 2t) \sigma^2]
        \leq & ~ \exp{(-t)}\\
        \Pr[k\sigma^2 - X \geq 2\sqrt{kt}\sigma^2]
        \leq & ~ \exp{(-t)}
    \end{align*}
    Further if $k \geq \Omega(\epsilon^{-2} t)$ and $t \geq \Omega(\log(1/\delta))$, then we have
    \begin{align*}
    \Pr[ | X - k \sigma^2 | \leq \epsilon k \sigma^2 ] \leq \delta.
    \end{align*}
\end{lemma}

\section{Problem Formulation}\label{sec:problem_formulation}
In previous formulation in \cite{dzps19,sy19}, they have normalization factor $\frac{1}{\sqrt{m}}$ and only work for ReLU activation function. In \cite{mosw22}, they don't have normalization factor and only work for ReLU activation function.

The statement refers to a specific type of neural network that has two layers. The hidden layer of the network consists of $m$ neurons, and the activation function  is the exponential function.
\begin{align*}
F (W,x,a) := \sum_{r=1}^m a_r \phi ( w_r^\top x ) ,
\end{align*}
To simplify optimization, we only focus on optimizing $W$ and not both $a$ and $W$ simultaneously, where $x \in \R^d$ represents the input, $w_1, \cdots, w_m \in \R^d$ are weight vectors in the first layer, and $a_1, \cdots, a_m \in \R$ are weights in the second layer.

$\forall r\in [m]$, given the function $\phi(x)=\exp(x)$,
we have
\begin{align}\label{eq:relu_derivative}
\frac{\partial F (W,x,a)}{\partial w_r}= a_r x\exp(\langle w_r,x\rangle).
\end{align}

\begin{definition}[$F(t)$, dynamic prediction]\label{def:u}
For any timestamp $t$, we define 
\begin{align*}
F_i(t) := \sum_{r=1}^m a_r \exp( \langle w_r(t) , x_i \rangle )
\end{align*}
\end{definition}

\begin{definition}[Loss function over time]
The objective function $L$ is defined as follows:
\begin{align*}
L (W(t) ) := \frac{1}{2} \sum_{i\in [n]} ( F_i(t)- y_i  )^2 .
\end{align*}
\end{definition}

Thus, we define
\begin{definition}[$\Delta w_r(t)$]\label{def:Delta_w_r_at_time_t}
We define $\Delta w_r(t) \in \R^d$, $\forall r \in [m]$ in the following:
\begin{align*}
\Delta w_r(t) : = \sum_{i=1}^n ( F_i(t) - y_i ) a_r x_i \exp( \langle w_r(t) , x_i \rangle ).
\end{align*}
\end{definition}

\begin{definition}[gradient descent update equation]\label{def:update}
The typical approach for optimizing the weight matrix $W$ involves applying the gradient descent algorithm in the following:
\begin{align*}
W(t+1) = W(t) - \eta \Delta W(t) .
\end{align*}
where $\Delta W(t) \in \R^{d \times m}$ and $\Delta w_r(t) \in \R^{d}$ is the $r$-th column of $\Delta W(t) \in \R^{d \times m}$.
\end{definition}

\newpage
\section{Initialization and Perturbation}\label{sec:intialization_and_perturbation}
In Section \ref{sec:intialization_and_perturbation}, we propose assumptions on initialization and analyze perturbations. In Section~\ref{sec:intialization_and_perturbation:tools}, some tools utilized in this paper are presented. In Section~\ref{sec:intialization_and_perturbation:diff_dsc_con}, we provide a bound on $\| H^{\dis} - H^{\cts} \|_F$ (the distinction between discrete and continuous) under the choice of $m$. In Section~\ref{sec:intialization_and_perturbation:bound_changes_H_w}, we demonstrate an upper bound on the difference between continuous and discrete under the assumption that $w$ is in a small ball. In Section~\ref{sec:intialization_and_perturbation:loss}, we show how to control the loss $\|y-F(0)\|_2^2=\|y\|_2^2$ at initialization by forcing $a_{2r} = -a_{2r-1}$.

\subsection{A list of tools}\label{sec:intialization_and_perturbation:tools}

Within this section, we demonstrate that the spectral proximity exists between the continuous and discrete versions of the gram matrix of input data.

\begin{definition}\label{def:B}
Let $C> 10$ denote a sufficiently large constant. 

We define parameter $B$ as follows 
\begin{align*}
 B:= C\sigma \sqrt{ \log(n/\delta) }.
\end{align*}
\end{definition}

\begin{lemma}\label{lem:bound_on_exp_w_and_perturb_w}
If the following conditions hold
\begin{itemize}
    \item Let $B > 0$ denote a parameter be defined as Definition~\ref{def:B}.
    \item Let $w_r$ denote random Gaussian vectors from ${\cal N}(0,\sigma^2 I_d)$.
    \item Let $v_r$ be the vector where $\| v_r - w_r \|_2 \leq R$, $\forall     r \in [m]$
    \item Let $x_i$ be the vector where $\| x_i \|_2 \leq 1$, $\forall i \in [n]$
    \item Let $R \in (0,0.01)$
\end{itemize}
Then, we have with probability $1-\delta$
\begin{itemize}
    \item Standard inner product
    \begin{itemize}
        \item Part 1. $| \langle w_r, x_i \rangle | \leq B$, $\forall i\in [n]$, $\forall r\in [m]$
        \item Part 2. $|\langle v_r, x_i \rangle | \leq B + R$, $\forall i\in [n]$, $\forall r\in [m]$
        \item Part 3. $| \langle w_r - v_r, x_i + x_j \rangle | \leq 2R$
    \end{itemize}
    \item $\exp$ function
    \begin{itemize}
        \item Part 4. $\exp(\langle w_r , x_i \rangle) \leq \exp( B )$, $\forall i\in [n]$, $\forall r\in [m]$
        \item Part 5. $\exp(\langle v_r, x_i \rangle) \leq \exp( B + R )$, $\forall i\in [n]$, $\forall r\in [m]$
        \item Part 6. $|\exp( \langle w_r - v_r, x_i + x_j \rangle ) - 1 | \leq 4R$
    \end{itemize}
\end{itemize}
\end{lemma}
\begin{proof}
{\bf Proof of Part 1, 2, 4 and 5.}

The proof is trivially follows from Gaussian tail bound.

{\bf Proof of Part 3 and 6.}

Because $x_i$ and $x_j$ are independent and $\|\Delta w_r\|_2\leq R$, we can have
\begin{align}\label{eq:x_i_x_j}
    |\langle \Delta w_r,(x_i+x_j)\rangle|\leq 2R \leq 0.1
\end{align}

Then, we have
\begin{align*}
    |\exp( \langle\Delta w_r, (x_i+x_j)\rangle)-1)|
    \leq & ~ 2 |\langle \Delta w_r, (x_i+x_j)\rangle| \notag \\
    \leq & ~ 4 R
\end{align*}
where the first step follows from Fact~\ref{fac:taylor}, and the last step follows from Eq.~\eqref{eq:x_i_x_j}.

\end{proof}

\subsection{Bounding changes between discrete and continuous}\label{sec:intialization_and_perturbation:diff_dsc_con}
In Section~\ref{sec:intialization_and_perturbation:diff_dsc_con}, we establish a bound on $\| H^{\dis} - H^{\cts} \|_F$ for the chosen value of $m = \Omega( \lambda^{-2} \cdot n^2 \cdot \exp(2B) \cdot \sqrt{\log(n/\delta)} )$. The following lemma can be viewed as a variation of Lemma 3.1 in \cite{dzps19}, a variation of Lemma 3.1 in \cite{sy19}, a variation of Lemma C.3 in \cite{bpsw21}, a variation of Lemma C.1 in \cite{syz21} a variation of Lemma G.1 in \cite{mosw22}.
\begin{lemma}[]\label{lem:3.1}

As per the definition, $H^{\cts}$ and $H^{\dis}$ are two matrices of size $n \times n$ that we specify as follows:
 
\begin{align*}
H^{\cts}_{i,j} := & ~ \E_{w \sim \N(0,I)} \left[ (\langle x_i,x_j\rangle)\cdot{ \exp(\langle w_r,x_i\rangle)\cdot\exp(\langle w_r,x_j\rangle) }\right] , \\ 
H^{\dis}_{i,j} := & ~ \frac{1}{m} \sum_{r\in [m]} \left[ (\langle x_i,x_j\rangle)\cdot \exp( \langle w_r,x_i\rangle) \cdot \exp(\langle w_r,x_j\rangle)\right].
\end{align*}
We define $\lambda :={  \lambda_{\min} (H^{\cts})} $. 

If the following conditions hold
\begin{itemize}
    \item $\lambda > 0$.
    \item $d =\Omega(\log(1/\delta))$.
    \item $m = \Omega( \lambda^{-2} \cdot n^2 \cdot \exp(2B) \cdot \sqrt{\log(n/\delta)} )$.
\end{itemize}
Then, we have

\begin{itemize}
    \item Part 1 $\| H^{\dis} - H^{\cts} \|_F \leq {  \frac{ \lambda }{4}}$.
    \item Part 2. $\lambda_{\min} ( H^{\dis} ) \geq {  \frac{3}{4} \lambda}$.
\end{itemize}
hold with probability at least $1-\delta$.
\end{lemma}

\begin{proof} 

{\bf Proof of Part 1.}
For any given pair $(i,j)$, $H_{i,j}^{\dis}$ is computed as the mean of a set of independent random variables, denoted as:
\begin{align*}
H_{i,j}^{\dis}=~\frac {1}{m}\sum_{r\in [m]} (\langle x_i, x_j\rangle)\cdot\exp(\langle w_r, x_i\rangle)\cdot \exp(\langle w_r, x_j\rangle).
\end{align*}

Then the expectation of $H_{i,j}^{\dis}$ is
\begin{align*}
\E [ H_{i,j}^{\dis} ]
= & ~\frac {1}{m}\sum_{r=1}^m \E_{w_r\sim {\N}(0, \sigma^2 I_d)} \left[ (\langle x_i, x_j\rangle)\cdot\exp(\langle w_r, x_i\rangle)\cdot\exp(\langle w_r,x_j\rangle)\right]\\
= & ~\E_{w\sim {\N}(0, \sigma^2 I_d)} \left[ (\langle x_i,x_j\rangle)\cdot\exp(\langle w, x_i\rangle)\cdot\exp(\langle w,x_j\rangle) \right]\\
= & ~ H_{i,j}^{\cts}.
\end{align*}
For each $r\in [m]$, we define $z_r$ as follows:
\begin{align}\label{eq:bound_for_wx}
z_r := \frac {1}{m}(x_i^\top x_j)\cdot\exp(w_r ^\top x_i)\cdot \exp(w_r^\top x_j).
\end{align}

Using Lemma~\ref{lem:bound_on_exp_w_and_perturb_w}, we have

Moreover, we have
\begin{align*}
|z_r| \leq \frac{1}{m} \exp( 2 B ) := M.
\end{align*}

So by Hoeffding inequality (Lemma~\ref{lem:hoeffding}) we have for all $t>0$,
\begin{align*}
\Pr \left[ | H_{i,j}^{\dis} - H_{i,j}^{\cts} | \geq t \right]
\leq & ~ {2\exp \Big( -\frac{2t^2}{4 M^2} \Big)} \\
 = & ~ { 2\exp(-t^2/(2 M^2))}.
\end{align*}
$t$ is chosen as follows:
\begin{align*} 
t= M   \sqrt{\log(n/\delta)} ,
\end{align*}

$\forall i,j \in [n]$, by applying the union bound to all pairs $(i,j)$, we can obtain that  with probability at least $1-\delta$ :
\begin{align*}
|H_{i,j}^{\cts}-H_{i,j}^{\dis}|
\leq & ~   M \sqrt{ \log(n/\delta) } \\
\leq & ~  \frac{1}{m} \exp(2 B ) \sqrt{\log(n/\delta)}.
\end{align*}
 
Therefore, we can conclude that:

\begin{align}\label{eq:lamda/4}
\|H^{\cts}-H^{\dis}\|_F^2 
 = & ~ \sum_{i\in [n]}\sum_{j\in [n]} |H_{i,j}^{\dis} - H_{i,j}^{\cts}|^2  \notag \\
 \leq & ~ \frac{1}{m} n^2 \exp(2B) \sqrt{\log(n/\delta)}  \notag \\
 \leq & ~ \lambda^2/16
\end{align}
The last step in the derivation follows directly from our choice of $m$.

{\bf Proof of Part 2}

Then we have
\begin{align*}
\lambda_{\min} ( H^{\dis} ) 
\geq & ~ \lambda_{\min} (H^{\cts}) - \|H^{\dis} - H^{\cts}\|  \notag \\
\geq & ~ \lambda_{\min} (H^{\cts}) - \|H^{\dis} - H^{\cts}\|_F  \notag \\
\geq & ~ \lambda - \lambda/4 \\
\geq & ~ 3\lambda/4
\end{align*}
 where the first step is due to Fact~\ref{fac:norm}, and the second step is from Fact~\ref{fac:norm}, the third step is from Eq.~\eqref{eq:lamda/4}, the fourth step is because of adding terms.
 \end{proof}

\subsection{Given \texorpdfstring{$w$}{}  within a small ball bounding changes of \texorpdfstring{$H$}{} }\label{sec:intialization_and_perturbation:bound_changes_H_w}
Under the assumption that $w$ is contained in a small ball, in Section~\ref{sec:intialization_and_perturbation:bound_changes_H_w}, we can restrict the discrepancy between the continuous and discrete versions 
\begin{definition}\label{def:delta_w}
    We define 
\begin{align*}
    \Delta w_r := \wt{w}_r-w_r
\end{align*}
and
\begin{align*}
    \|\Delta w_r\|_2\leq R
\end{align*}
\end{definition}

\begin{definition}\label{def:c}
    We define 
\begin{align*}
    z_i:=\wt{w}_r^\top x_i
\end{align*}
\end{definition}
\begin{definition}\label{def:s_r}
   For $i\in [n],j\in [n]$, $r\in [m]$, we define
\begin{align*}
s_{r,i,j} := \exp({ \wt{w}_r^\top x_i})\cdot \exp( \wt{w}_r^\top x_j)  - \exp({ w_r^\top x_i})\cdot \exp( w_r^\top x_j).
\end{align*} 
By fixing $i$ and $j$, $s_{r,i,j}$ is simplified to $s_r$ with $s_r$ as a random variable that depends solely on $\wt{w}_r$. The set of random variables ${s_r}_{r=1}^m$ are independent of each other, because $\{\wt{w}_r\}_{r=1}^m$ are independent. 
\end{definition}
The following Lemma can be viewed as a variation of Lemma 3.2 in \cite{sy19}, a variation of Lemma C.4 and C.5 in \cite{bpsw21} a variation Lemma C.2 in \cite{syz21} and a variation of Lemma G.2 in \cite{mosw22}.
\begin{lemma}[perturbed $w$]\label{lem:perturb_w}
Suppose $\wt{w}_1, \cdots, \wt{w}_m$ are independent and identically distributed with a normal distribution ${\N}(0,\sigma^2 I)$, and let $B$ be defined as in Definition~\ref{def:B}. For any set of weight vectors $w_1, \cdots, w_m \in \R^d$ that satisfy $\| \wt{w}_r - w_r \|_2 \leq R$  where $R\in (0,0.001)$, for any $r\in [m]$, we define the function $H : \R^{m \times d} \rightarrow \R^{n \times n}$ as follows
\begin{align*}
    H(w)_{i,j} =  \frac{1}{m} x_i^\top x_j \sum_{r\in [m]} \exp({ w_r^\top x_i})\cdot \exp( w_r^\top x_j) .
\end{align*}
Therefore we can conclude that with probability at least $1-(n^2 \cdot \exp(-m R/10)+\delta)$, we have
\begin{align*}
\| H (w) - H(\wt{w}) \|_F \leq 3nR \exp(2B),
\end{align*}
\end{lemma}

\begin{proof}

The random variable we care is

\begin{align*}
& ~ \sum_{i\in [n]} \sum_{j\in [n]}| H(\wt{w})_{i,j} - H(w)_{i,j} |^2 \\
\leq & ~ \frac{1}{m^2} \sum_{i\in [n]} \sum_{j\in [n]} \left( \sum_{r\in [m]} \exp({ \langle \wt{w}_r, x_i\rangle})\cdot \exp( \langle \wt{w}_r, x_j\rangle)  - \exp({ \langle w_r, x_i\rangle})\cdot \exp( \langle w_r, x_j\rangle) \right)^2 \\
= & ~ \frac{1}{m^2} \sum_{i\in [n]} \sum_{j\in [n]}  \Big( \sum_{r\in [m]} s_{r,i,j} \Big)^2 ,
\end{align*}

where the last step is due to $\forall r,i,j$.

For simplicity, we drop the index of $i,j$ in $s_{r,i,j}$. We only keep the $r$, i.e., using $s_r$ to denote $s_{r,i,j}$.

We define $s_r$ as follows
\begin{align*}
s_r : = \exp({ \wt{w}_r^\top x_i})\cdot \exp( \wt{w}_r^\top x_j)  - \exp({ w_r^\top x_i})\cdot \exp( w_r^\top x_j) .
\end{align*}

Using Lemma~\ref{lem:bound_on_exp_w_and_perturb_w}, we have
\begin{align}\label{eq:bound_s_r}
\Pr [\forall r \in [m], s_r \leq \exp(2B) ] \geq 1- \delta.
\end{align}
In the rest of the proof, we will condition on the above event is holding.

We can obtain the upper bound of $\E_{\wt{w}_r}[s_r]$, we have
\begin{align}\label{eq:rewrite_E_s_r}
    \E_{ \wt{w}_r }[ s_r ]
    \leq & ~ \E_{\wt{w}_r}[|{\exp({ \wt{w}_r^\top x_i})\cdot \exp( \wt{w}_r^\top x_j)-\exp({ w_r^\top x_i})\cdot \exp( w_r^\top x_j)}|] \notag \\
    \leq & ~ \E_{\wt{w}_r}[|{\exp({ {(w_r+\Delta w_r)}^\top x_i})\cdot \exp( {(w_r+\Delta w_r)}^\top x_j)-\exp({ w_r^\top x_i})\cdot \exp( w_r^\top x_j)}|] \notag \\
     \leq & ~ \E_{\wt{w}_r}[|(\exp(\Delta w_r^\top (x_i+x_j))-1)\cdot\exp({ w_r^\top x_i})\cdot \exp( w_r^\top x_j)] \notag \\
     \leq & ~ \exp({ w_r^\top x_i})\cdot \exp( w_r^\top x_j) \cdot \E_{\wt{w}_r}[|(\exp(\Delta w_r^\top (x_i+x_j))-1)|]
\end{align}
where the 1st step is due to Definition~\ref{def:s_r}, the 2nd step is because of Definition~\ref{def:delta_w}, the 3rd step is by selecting the same term $\exp( w_r^\top x_i )\cdot \exp( w_r^\top x_j )$, and the last step follows from the reason that $i,j$ are fixed.

We have
\begin{align}\label{eq:bound_E_s_r}
     \E_{\wt{w}_r}[s_r]
     \leq & ~ \exp({ w_r^\top x_i})\cdot \exp( w_r^\top x_j) \cdot \E_{\wt{w}_r}[|(\exp(\Delta w_r^\top (x_i+x_j))-1)|] \notag \\
     \leq & ~ \exp(2 B) \cdot \E_{\wt{w}_r}[|\exp(\Delta w_r^\top (x_i+x_j)) - 1|] \notag \\
     \leq & ~ \exp(2B) \cdot \E[4R] \notag \\
    \leq & ~ 4R\cdot \exp(2B)
\end{align}
where the 1st step is due to Eq.~\eqref{eq:rewrite_E_s_r}, the 2nd step is due to Lemma~\ref{lem:bound_on_exp_w_and_perturb_w}, and the 3rd step is due to Lemma~\ref{lem:bound_on_exp_w_and_perturb_w}, and the last step follows from simple algebra.

We have, 
\begin{align}\label{eq:bound_for_s_r_2}
    \E_{\wt{w}_r} \left[ \left( s_r -\E_{\wt{w}_r} [s_r] \right)^2 \right]
    = & ~ \E_{\wt{w}_r} [s_r^2]- (\E_{\wt{w}_r} [s_r])^2  \notag \\
    \leq & ~ \E_{\wt{w}_r}[s_r^2] \notag \\
    = & ~\E_{\wt{w}_r} \left[(\exp({ \wt{w}_r^\top x_i})\cdot \exp( \wt{w}_r^\top x_j)-\exp({ w_r^\top x_i})\cdot \exp( w_r^\top x_j))^2\right] \notag \\
    \leq & ~ \E_{\wt{w}_r} \left[( \exp(4 B ) (\exp(\Delta w_r^\top (x_i + x_j))-1))^2\right] \notag \\
    \leq & ~ 16 R^2 \cdot \exp(4B)
\end{align}
where the 1st step is due to the definition of variance, the 2nd step is due to simple algebra, the 3rd step is due to the Definition~\ref{def:s_r}, the 4th step follows from Lemma~\ref{lem:bound_on_exp_w_and_perturb_w}, the 5th step follows from Lemma~\ref{lem:bound_on_exp_w_and_perturb_w}.

In the rest of the proof, let us condition on the above event is holding.

Combining Eq.~\eqref{eq:bound_s_r} Eq.~\eqref{eq:bound_E_s_r}, we also have
\begin{align*}
|s_r-\E_{\wt{w}_r}[s_r]|
\leq & ~ (1+4R)\cdot \exp(2B)\\
\leq & ~ 2 \exp(2B),
\end{align*}
 where the second step is from $4R\leq 1$.

Using Bernstein inequality (Lemma~\ref{lem:bernstein}), we have
\begin{align*}
\Pr \left[ Z > t \right] \leq \exp \left( - \frac{ t^2/2 }{ \Var[Z]  + M t /3 } \right).
\end{align*}
where 
\begin{align*}
    Z: = & ~ \sum_{r=1}^m s_r - \E[s_r], \\
    \Var[Z] := & ~ 16 m R^2 \exp(4 B ), \\
    M := & ~ 2 \exp(2 B).
\end{align*}

Replacing $t =  m R \exp(2B)$, we know that
\begin{align*}
\Var[Z] + M t/ 3
= & ~ 16 m R^2 \exp(4B) + 2 \exp(2B) m R \exp(2B) / 3 \\
\leq & ~ m R \exp(4B)
\end{align*}
Thus, 
\begin{align*}
\frac{t^2/2}{\Var[Z] + M t/ 3} \geq  \frac{ (m R \exp(B))^2 }{ m R \exp(4B)  } = m R 
\end{align*}
and 
\begin{align}\label{eq:bound_pr_sr}
    \Pr \left[ X \geq  m R \exp(2B) \right]
    \leq & ~ \exp \left( - m R \right) 
\end{align}

Using the bound on expectation
\begin{align*}
\Pr \left[\frac{\sum_{r\in [m]}s_r}{m} \geq 3R \cdot \exp(2B)  \right] 
   \leq &~\exp \left( - Rm/10 \right) \\
\end{align*}

\end{proof}

\subsection{Controlling the Loss at initialization}\label{sec:intialization_and_perturbation:loss}

The initialization of choice $a \in \{-1,+1\}^m$ is purely random in \cite{dzps19,sy19}. That initialization will make $\|  F(0) \|_2^2$ is a bit large. We can use the initialization idea in \cite{mosw22}. The idea is forcing $a_{2r} = -a_{2r-1}$. This will gives us that $\| F(0)\|_2^2= 0$.
 \begin{claim}\label{cla:yu0}
We have
\begin{align*}
\|y-F(0)\|_2^2= \| y \|_2^2.
\end{align*}
\end{claim}
\begin{proof}
We also duplicate the weights such that $w_{2r} = w_{2r-1}$. Therefore, the proof is straightforward.
\end{proof}

\section{Convergence}\label{sec:convergence}
In Section~\ref{sec:convergence}, when the neural network is excessively over-parametrized, we observe a linear decrease in the training error, leading to its ultimate convergence to 0.
In Section~\ref{sec:converge:mainresult}, we present our paper's main result. Section~\ref{sec:converge:induction_weights} outlines the induction lemma for weights, while Section~\ref{sec:converge:induction_loss} introduces the induction lemma for loss. Finally, in Section~\ref{sec:coverge:induction_gradient}, we provide the induction lemma for gradient.

\subsection{Main Result}\label{sec:converge:mainresult}

\begin{theorem}[Main result, formal version of Theorem~\ref{thm:informal}]\label{thm:formal}
If the following conditions hold
\begin{itemize}
    \item Let  $\lambda=\lambda_{\min}(H^{\cts})>0$
    \item  $m = \Omega( \lambda^{-2} n^2 \exp(4 B) \log^2(n/\delta) )$
    \item Let $w_r$ and $a_r$ be defined as Definition~\ref{def:duplicate_weights}.
    \item Let $\eta=0.01 \lambda / ( m n^2 \exp(4 B) )$
    \item Let $T = \Omega( (m \eta \lambda)^{-1} \log(n/\epsilon)  ) = \Omega( \lambda^{-2} n^2 \exp(4B) \cdot \log(n/\epsilon) )$
\end{itemize}
Then, we have running algorithm with $T$ iterations
\begin{align*}
    \| F(T) - y \|_2^2 \leq \epsilon
\end{align*}
\end{theorem}
\begin{proof}

We choose $\sigma = 1$.

We have proved $\| F(0) - y \|_2^2 \leq n$.

Using the choice of $T$, then it directly follows from alternatively applying Lemma~\ref{lem:induction_part_2_loss} and Lemma~\ref{lem:induction_part_1_weights}.

Since $\exp(\Theta(B)) = n^{o(1)}$, then in Theorem~\ref{thm:informal}, we can simplify the $n^2 \exp(\Theta(B)) = n^{2+o(1)}$.
\end{proof}

\subsection{Induction Part 1. For Weights}\label{sec:converge:induction_weights}

\begin{lemma}[Induction Part 1 for weights]\label{lem:induction_part_1_weights}
If the following condition hold
\begin{itemize}
    \item General Condition 1. Let $\lambda = \lambda_{\min} (H^{\cts}) > 0$
    \item General Condition 2. $\eta=0.01\lambda /(mn^2 \exp{(4 B)})$
    \item General Condition 3. Let $D$ be defined as Definition~\ref{def:D}
    \item General Condition 4. Let $w_r$ and $a_r$ be defined as Definition~\ref{def:duplicate_weights}.
    \item General Condition 5. $D < R$
    
    \item {\bf Weights Condition.} $\| w_r(i) - w_r(0)\|_2 \leq  R$ for all $i \in [t]$
    \item {\bf Loss Condition.}  $\| F(i) - y \|_2^2 \leq \| F(0) - y \|_2^2\cdot (1-m\eta \lambda/2)^i$,  $\forall i \in [t]$
    \item {\bf Gradient Condition.} $\eta \| \Delta w_r(i) \|_2 \leq 0.01$ for all $r \in [m]$, for all $i
    \in [t]$
\end{itemize}

For $t+1$ and $\forall r\in [m]$, it holds that:
\begin{align*}
\| w_r(t+1) - w_r(0) \|_2 \leq D.
\end{align*}
\end{lemma}

\begin{proof}

We have
\begin{align} \label{eq:upper_bound_etasum}
& ~ \eta \sum_{i=0}^\infty (1-n\lambda/2)^{i/2}\notag\\
= & ~ \eta \sum_{i=0}^\infty (1-\eta \lambda/4)^i \notag \\
\leq & ~ \eta \frac{1}{ \eta \lambda / 4 } \notag \\
\leq & ~ 8/\lambda
\end{align}
where the 1st step is due to Fact~\ref{fac:taylor}, the 2nd step is due to Fact~\ref{fac:taylor}, the last step is due to simple algebra.

Our approach involves utilizing the gradient's norm as a means of constraining the distance as follows:
\begin{align*}
& ~ \|w_r(0)-w_r(t+1)\|_2\\
\le & ~\eta \sum_{i=0}^{t} \| \Delta w_r(i) \|_2 \\
\le & ~ \eta \sum_{i=0}^{t} \exp(B+R) \cdot \sqrt{n}  \cdot \|F(i)-y\|_2 \\
\le & ~ \eta \sum_{i=0}^{t} (1-{\eta \lambda}/{2})^{i/2} \cdot \exp(B +  R) \cdot \sqrt{n} \cdot \|F(0)-y\|_2 \\
\le & ~ 8 \sqrt{n} \cdot \lambda^{-1} \cdot \exp(B + R)  \|F(0)-y\|_2 \\
= & ~ D
\end{align*}
where the 1st is from $w_r(s+1)-w_r(s)=\eta \cdot \Delta w_r(s)$, the 2nd step is due to Lemma~\ref{lem:bound_Delta_w_at_time_s} for $m t$ times, the 3rd step is due to Condition 3 in Lemma statement, the forth step is due to simple algebra, and the forth step is due to Eq.~\eqref{eq:upper_bound_etasum}, the last step is due to Condition 2 in Lemma statement.
\end{proof}

\subsection{Induction Part 2. For Loss}\label{sec:converge:induction_loss}
Now, we present our next induction lemma.
\begin{lemma}[Induction Part 2. For Loss]\label{lem:induction_part_2_loss}
Let $t$ be a fixed integer. 

If the following conditions hold
\begin{itemize}
   \item General Condition 1. Let $\lambda = \lambda_{\min} (H^{\cts}) > 0$
    \item General Condition 2. $\eta=0.01\lambda /(mn^2 \exp{(4 B)})$
    \item General Condition 3. Let $D$ be defined as Definition~\ref{def:D}
    \item General Condition 4. Let $w_r$ and $a_r$ be defined as Definition~\ref{def:duplicate_weights}.
    \item General Condition 5. $D < R$
    \item {\bf Weight Condition.} $\| w_r(t) - w_r(0) \|_2 \leq D < R$, $\forall r \in [m]$
    \item {\bf Loss Condition.} $\| F(i) - y \|_2^2 \leq (1-m\eta \lambda/2)^i \cdot \| F(0) - y \|_2^2$, for all $i \in [t]$
    \item {\bf Gradient Condition.} $\eta \| \Delta w_r(i) \|_2 \leq 0.01$  $\forall r \in [m]$, $\forall i\in [t]$
\end{itemize}
Then we have
\begin{align*}
\| F (t+1) - y \|_2^2 \leq ( 1 - m \eta \lambda / 2 )^{t+1} \cdot \| F (0) - y \|_2^2.
\end{align*}
\end{lemma}
\begin{proof}

Recall the update rule (Definition~\ref{def:update}),
\begin{align*}
w_{r}(t+1) = w_r(t) - \eta \cdot \Delta w_{r}(t)
\end{align*}

$\forall i \in [n]$, it follows that

\begin{align*}
& ~ F_i(t+1) - F_i(t) \\
= & ~   \sum_{r\in [m]} a_r \cdot ( \exp( \langle w_r(t+1),x_i \rangle) - \exp(\langle w_r(t),x_i \rangle) )  \\
= & ~  \sum_{r\in [m]} a_r \cdot \exp(\langle w_r(t),x_i\rangle) \cdot ( \exp(- \eta \langle \Delta w_r(t),x_i\rangle) - 1 ) \\
= & ~  \sum_{r\in [m]} a_r \cdot \exp(w_r(t)^\top x_i) \cdot (-\eta \langle \Delta w_r(t), x_i \rangle + \Theta(1) \eta^2 \langle \Delta w_r(t), x_i \rangle^2 ) \\
= & ~ v_{1,i} + v_{2,i}
\end{align*}
where the third step follows from $|\eta \Delta w_r(t)^\top x_i| \leq 0.01$ and Fact~\ref{fac:taylor},  the last step is from
\begin{align*}
v_{1,i}:= & ~ \sum_{r=1}^m a_r \cdot \exp(\langle w_r(t),x_i\rangle) \cdot (-\eta \langle \Delta w_r(t), x_i \rangle ) \\
v_{2,i}:= & ~ \sum_{r=1}^m a_r \cdot \exp(\langle w_r(t),x_i\rangle) \cdot \Theta(1) \cdot \eta^2 \cdot  \langle \Delta w_r(t), x_i \rangle^2
\end{align*}
Here $v_{1,i}$ is linear in $\eta$ and $v_{2,i}$ is quadratic in $\eta$. Thus, $v_{1,i}$ is the first order term, and $v_{2,i}$ is the second order term.

Recall the definition of $H$ over timestamp $t$ (see Definition~\ref{def:H_s})
\begin{align*}
H(t)_{i,j} = & ~ \frac{1}{m} \sum_{r\in [m]} x_i^\top x_j \exp( \langle w_r(t),x_i\rangle) \cdot \exp( \langle w_r(t),x_j\rangle) ,  
\end{align*}
Further, we define $C_1, C_2, C_3$
\begin{align*}
C_1 = & ~ -2 \eta (F(t)- y )^\top H(t) ( F(t)-y) , \\ 
C_2 = & ~ - 2 (F(t)- y  )^\top v_2 , \\
C_3 = & ~ \| F (t+1) - F(t) \|_2^2 . 
\end{align*}
Then we can rewrite
\begin{align*}
\| y -F(t+1) \|_2^2 = \| y - F(t) \|_2^2 + C_1 + C_2 + C_3
\end{align*}

We have
\begin{align*}
  \|F(t)-y\|_2^2 
 \leq  \|F(t-1)-y\|_2^2 \cdot (1-m\eta \lambda/2) 
\end{align*}
 where the first step follows is due to Lemma~\ref{lem:loss_one_step_shrinking}.

Thus, we complete the proof.
\end{proof}

\subsection{Induction Part 3. For Gradient }\label{sec:coverge:induction_gradient}

\begin{lemma}[Induction Part 3. For Loss]\label{lem:induction_part_3_gradient}
Let $t$ be a fixed integer. 

If the following conditions hold
\begin{itemize}
   \item General Condition 1. Let $\lambda = \lambda_{\min} (H^{\cts}) > 0$
    \item General Condition 2. $\eta=0.01\lambda /(mn^2 \exp{(4 B)})$
    \item General Condition 3. Let $D$ be defined as Definition~\ref{def:D}
    \item General Condition 4. Let $w_r$ and $a_r$ be defined as Definition~\ref{def:duplicate_weights}.
    \item General Condition 5. $D < R$
    \item {\bf Weight Condition.} $\| w_r(t) - w_r(0) \|_2 \leq D < R$, $\forall r \in [m]$
    \item {\bf Loss Condition.} $\|  F(i) - y \|_2^2 \leq \| F(0) - y \|_2^2 \cdot (1-m\eta \lambda/2)^i $, $\forall i \in [t]$
    \item {\bf Gradient Condition.} $\eta \| \Delta w_r(i) \|_2 \leq 0.01$  $\forall r \in [m]$, $\forall i \in [t]$
\end{itemize}
Then we have
\begin{align*}
\eta \| \Delta w_r(t+1) \|_2 \leq 0.01, \forall r \in [m]
\end{align*}
\end{lemma}
\begin{proof}

We have
\begin{align*}
\eta \| \Delta w_r(t+1) \|_2 
= & ~ \eta \left\| \sum_{i=1}^n  a_r x_i \cdot (y_i - F_i(t+1))   \cdot \exp(  \langle w_r(t+1), x_i\rangle ) \right\|_2 \notag\\
\leq & ~ \eta \exp (B + R) \cdot    \sum_{i=1}^n | y_i - F_i(t+1) | \notag\\
\leq & ~ \eta \exp (B + R)  \cdot  \sqrt{n} \cdot \| y - F(s) \|_2  \\
\leq & ~ \eta \exp (B + R)  \cdot  \sqrt{n} \cdot \| y - F(0) \|_2  \\
\leq & ~ \eta \exp (B + R)  \cdot n \\
\leq & ~ 0.01
\end{align*}
where the 1st step follows from Definition~\ref{def:Delta_w_r_at_time_t}, the 2nd step is due to Lemma~\ref{lem:bound_on_exp_w_and_perturb_w}, the 3rd step is due to Cauchy-Schwartz inequality, the 4th step follows is due to {\bf Loss Condition}, the 5th step follows from $\| y-F(0) \|_2 = \sqrt{n}$, the sixth step is due to the choice of $\eta$.

\end{proof}

\newpage
\section{Induction Part 1: For Weight}\label{sec:induction_for_weight}
In Section~\ref{sec:induction_for_weight}, we present the weight bound, which helps us complete the proof. Section~\ref{sec:induction:definition} introduces various definitions used throughout the paper, while Section~\ref{sec:induction:bound_gradient} proposes the bounding gradient lemma and its corresponding proof.

\subsection{Definition of \texorpdfstring{$D$}{}}\label{sec:induction:definition}

To simplify the notation, we present the definition as follows.
\begin{definition}\label{def:D}
We define $D_{\cts}$ 
\begin{align*}
D := 8 \cdot \lambda^{-1} \cdot \exp( B + R ) \cdot \frac{ \sqrt{n} }{ m } \cdot \| y - F(0) \|_2 .
\end{align*}
\end{definition}

We define the kernel with respect to timestample $s$.
\begin{definition}\label{def:H_s}
Let $H(s) \in \R^{n \times n}$ be a matrix defined for any $s$ in the interval $[0,t]$.
\begin{align*}
H(s)_{i,j} := \frac{1}{m} \sum_{r\in [m]} x_i^\top x_j \cdot \exp{(\langle w_r(s),x_i\rangle)} \cdot \exp{(\langle w_r(s),x_j\rangle)}.
\end{align*} 
\end{definition}
\begin{definition} \label{def:H_asy}
For any matrix $P \in [-1,1]^{m \times n}$, we define asymmetric matrix $H_{\asy}$

\begin{align*}
H_{\asy}(s)_{i,j} := \frac{1}{m} \sum_{r\in [m]} x_i^\top x_j \cdot p_{i,r} \cdot \exp{(\langle w_r(s),x_i\rangle)} \cdot \exp{(\langle w_r(s),x_j\rangle)}.
\end{align*} 

\end{definition}
\begin{claim} \label{cla:upper_bound_HP}
We have
\begin{align*}
\| H_{\asy} (s) \|_{\infty} \leq \exp( 2 (B + R) ).
\end{align*}
holds with probability $1-\delta$.
\end{claim}
\begin{proof}

\begin{align*}
     \|H(P)\|_\infty
=  \max_{i\in [n], j\in [n]}\{ \frac{1}{m} \sum_{r\in [m]} x_i^\top x_j \cdot p_{i,r} \cdot \exp{(w_r(s)^\top x_i)} \cdot \exp{(w_r(s)^\top x_j)} \} 
\end{align*}
where the first step is from Definition~\ref{def:H_asy}.

It is sufficient to make a bound for each $i \in [n]$ and $j \in [n]$.

We have
\begin{align*}
  & ~ \frac{1}{m} \sum_{r\in [m]} x_i^\top x_j \cdot p_{i,r} \cdot \exp{(\langle w_r(s), x_i\rangle)} \cdot \exp{(\langle w_r(s), x_j\rangle)} \} \\
 \leq ~ &  \frac{1}{m} \sum_{r\in [m]}  \exp{(\langle w_r(s),x_i\rangle)} \cdot \exp{(\langle w_r(s), x_j\rangle)}\\
\leq ~ &  \frac{1}{m} \sum_{r\in [m]}  \exp{(2 (R+B))}\\
= ~ & \exp(2(B+R))
\end{align*}
 the 1st step is from $\| x_i \|_2 \leq 1$ and $|p_{i,r}| \leq 1$, the second step is due to Lemma~\ref{lem:bound_on_exp_w_and_perturb_w}.
\end{proof}

\subsection{Bounding the gradient at any time}\label{sec:induction:bound_gradient}
In this section, we bound the gradient at any time.
\begin{lemma}\label{lem:bound_Delta_w_at_time_s}
It the following condition hold,
\begin{itemize}
    \item $\| w_r(s) - w_r(0) \|_2 \leq R$
\end{itemize}
For any timestamp at time $s$, we have
\begin{align*}
\| \Delta w_r(s) \|_2 
\leq \exp(B+R )  \sqrt{n}  \| y - F(s) \|_2
\end{align*}
\end{lemma}
\begin{proof}

We have
\begin{align*}
\| \Delta w_r(s) \|_2 
= & ~  \left\| \sum_{i=1}^n (y_i - F_i)   a_r x_i \cdot \exp(  w_r(s)^\top x_i ) \right\|_2 \notag\\
\leq & ~ \exp (B + R) \cdot    \sum_{i=1}^n | y_i - F_i(s) | \notag\\
\leq & ~ \exp (B + R)  \cdot  \sqrt{n} \cdot \| y - F(s) \|_2 
\end{align*}
where the first step follows from Definition~\ref{def:Delta_w_r_at_time_t}, the second step follows from Lemma~\ref{lem:bound_on_exp_w_and_perturb_w}, the third step follows from Cauchy-Schwartz inequality.
\end{proof}

\begin{lemma}\label{lem:bound_Delta_w_times_eta}
It the following condition hold,
\begin{itemize}
    \item $\eta = 0.01 \lambda /(m n^2 \exp(4B))$
    \item $\| w_r(s) - w_r(0) \|_2 \leq R$
\end{itemize}
For any timestamp at time $s$, we have
\begin{align*}
\eta \| \Delta w_r(s) \|_2 
\leq 0.01
\end{align*}
\end{lemma}
\begin{proof}
This trivially follows from choice of $\eta$.
\end{proof}

\newpage
\section{Induction Part 2: For Loss}\label{sec:induction_for_loss}

In Section~\ref{sec:induction_for_loss}, we establish a bound for the loss at any time. To accomplish this, we decompose the loss $\|y-F(k+1)\|_2^2$ into three parts, namely $C_1,C_2$, and $C_3$, which are defined and discussed in Section~\ref{sec:indcution_for_loss:decomposition}. We provide our choices for $m$ and $\eta$ in Section~\ref{sec:induction_for_loss:choice_parameters}, while Section~\ref{sec:induction_for_loss:C_1}, Section~\ref{sec:induction_for_loss:C_2}, and Section~\ref{sec:induction_for_loss:C_3} respectively establish bounds for $C_1,C_2$, and $C_3$.

\subsection{Decomposition for \texorpdfstring{$\|y-F(t+1)\|_2^2$}{}}\label{sec:indcution_for_loss:decomposition}

In this section, we decompose the loss $\|y-F(t+1)\|_2^2$ into three parts $C_1, C_2$ and $C_3$.

\begin{lemma}\label{lem:rewrite_shrinking_one_step}
Assuming the following condition is met:
\begin{itemize}
    \item $C_1 = -2m \eta (F(t)-y)^\top H(t) (F(t)-y)$
    \item $C_2 = 2 m \eta^2 (F(t)-y)^\top H_{\asy} (t) (F(t)-y)$
    \item $C_3 = \| F(t+1) - F(t) \|_2^2$
\end{itemize}
then
\begin{align*}
\|F(t+1) - y\|_2^2 \leq \| F(t )-y\|_2^2 + C_1 + C_2 + C_3
\end{align*}
\end{lemma}
\begin{proof}
In the following manner, we can express $F(t+1) - F(t) \in \R^n$:
\begin{align*}
F(t+1) - F(t) = v_1 + v_2 .
\end{align*}

Using the notation of $H$, we can express $v_{1,i} \in \R$ as follows: 

\begin{align*}
v_{1,i} 
= & ~  \sum_{r\in [m]} a_r  \cdot \exp(\langle x_i,w_r(t)\rangle) \cdot (-\eta \langle x_i,\Delta w_r(t) \rangle ) \\
= & ~  \sum_{r\in [m]} a_r  \cdot \exp(\langle x_i,w_r(t)\rangle) \cdot ( - \eta  \sum_{j\in [n]} (F_j(t) - y_j) a_r x_j^\top \exp( w_r^\top x_j ) ) x_i\\
= & ~ - m\eta \cdot \frac{1}{m} \sum_{j\in [n]} x_i^\top x_j (F_j - y_j) \sum_{r\in [m]} \exp( \langle w_r(t) , x_i \rangle ) \cdot \exp( \langle w_r(t) , x_j \rangle) \\
= & ~ - m \eta \cdot  \sum_{j\in [n]} (F_j - y_j) ( H_{i,j}(t) ) ,
\end{align*}
where the second step follows from Definition~\ref{def:Delta_w_r_at_time_t}.

The equation above indicates that the vector $v_1 \in \R^n$ can be expressed as
\begin{align}\label{eq:rewrite_v1}
v_1 = m \eta ( y - F(t) )^\top ( H( t ) ) .
\end{align}

Let $p_{i,r} \in [-1,1]$.
Similarly
\begin{align*}
v_{2,i} 
= & ~ \sum_{r\in [m ]} a_r  \cdot \exp(\langle w_r(t),x_i\rangle) \cdot (\eta^2 p_{i,r} \langle \Delta x_i,w_r(t) \rangle ) \\
= & ~ \sum_{r\in [m]} a_r  \cdot \exp(\langle w_r(t),x_i\rangle) \cdot (  \eta^2 p_{i,r} \cdot \sum_{j=1}^n (F_j(t) - y_j) a_r x_j^\top \exp( \langle w_r, x_j \rangle) ) x_i\\
= & ~  m \eta^2 \frac{1}{ m } \sum_{j\in [n]} x_i^\top x_j (F_j - y_j) \sum_{r=1}^m p_{i,r} \exp( \langle w_r(t) , x_i \rangle ) \cdot \exp( \langle w_r(t) , x_j \rangle) \\
= & ~ m \eta^2 \sum_{j=1}^n (F_j - y_j) ( (H_{\asy}(t))_{i,j} ) ,
\end{align*}
The expression $\| y - F(t+1) \|_2^2$ can be rewritten in the following:
\begin{align*}
& ~\| y - F(t+1) \|_2^2\\
= & ~ \| y - F(t) - ( F(t+1) - F(t) ) \|_2^2 \\
= & ~ \| y - F(t) \|_2^2 - 2 ( y - F(t) )^\top  ( F(t+1) - F(t) ) +  \| F(t+1) - F(t) \|_2^2 .
\end{align*}

We can rephrase the second term in the Equation above as follows:
\begin{align*}
 & ~ \langle y - F(t), F(t+1) - F(t)\rangle \\
= & ~ \langle y - F(t) , v_1 + v_2 \rangle \\
= & ~ \langle y - F(t) , v_1\rangle + \langle y - F(t),v_2\rangle  \\
= & ~ m\eta (  F(t)-y )^\top H(t) (F(t) -y) - m \eta^2 ( F(t)-y )^\top H_{\asy}  (F(t)-y),
\end{align*}
where the third step is from Eq.~\eqref{eq:rewrite_v1}.

Therefore, we can conclude that
\begin{align*}
\| F(t+1)-y \|_2^2 
\leq  \| F(t)- y \|_2^2 + C_1 + C_2 + C_3
\end{align*}
where the last step follows from Lemma~\ref{lem:loss_one_step_shrinking}. 
\end{proof}

\subsection{Choice of Parameters}\label{sec:induction_for_loss:choice_parameters}
In  this section, we propose our choice of parameters $m,\eta, R, B$.
\begin{lemma} \label{lem:loss_one_step_shrinking}
    If the following conditions hold
\begin{itemize}
    \item Condition 1. $m = \Omega( \lambda^{-2} n^2 \exp(4 B) \log(n/\delta))$
    \item Condition 2. $\eta=0.01 \lambda / ( m n^2 \exp(4 B) ) $ 
    \item Condition 3. $R = 0.01 \lambda / ( n \exp( B) )$  
    \begin{itemize} 
        \item Required by Claim~\ref{cla:C1}
    \end{itemize}
    \item Condition 4. $R \leq 1 \leq B$
    \begin{itemize}
        \item Required by Claim~\ref{cla:C2} and Claim~\ref{cla:C3}
    \end{itemize}
    \item Condition 5. $D = 8 \lambda^{-1} \exp(B+R) \frac{ \sqrt{n} }{m} \| y - F(0) \|_2$
    \item Condition 6. $D < R$
    
    \item Condition 7. $\eta \| \Delta_r (t) \|_2 \leq 0.01$ for all $r \in [m]$
    \begin{itemize}
        \item Required by Claim~\ref{cla:C3}
    \end{itemize}
\end{itemize}
Then it holds that
\begin{align*}
    \|F(t+1)- y  \|_2^2 \leq  \|  F(t)-y  \|_2^2 \cdot ( 1 - m \eta \lambda / 2 )
\end{align*}
holds with probability $1-\delta$.
\end{lemma}
\begin{proof}
We can show
\begin{align*}
\| F(t+1)-y \|_2^2 
\leq & ~ \|  F(t)-y \|_2^2 + C_1 + C_2 + C_3 \\
\leq & ~ (1- m \eta \lambda + 2 m^2 \eta^2 n^2 \exp(4 B)) \cdot \|  F(t)-y \|_2^2 .
\end{align*}
where the first step follows from Lemma~\ref{lem:rewrite_shrinking_one_step}, the second step follows from Claim~\ref{cla:C1}, Claim~\ref{cla:C2}, and Claim~\ref{cla:C3}.

\paragraph{Choice of $\eta$.}

Next, we want to choose $\eta$  such that
\begin{equation}\label{eq:choice_of_eta_R}
(1- m \eta \lambda + 2 m^2 \eta^2 n^2 \exp(4 B))\leq (1- m \eta\lambda/2) .
\end{equation}

Using the choice of $\eta$ in Condition 2
\begin{align*}
 2 m^2 \eta^2 n^2 \exp(4 B) \leq  m \eta \lambda /4 
\end{align*}
This indicates:
\begin{align}\label{eq:1-eta_lambda_2}
\|  F(t+1) -y\|_2^2 \leq & ~( 1 - m \eta \lambda / 2 )\cdot \|F(t)-y\|_2^2  
\end{align}

\paragraph{lower bound for $m$, over-parametrization Size}

We require the following two conditions
\begin{itemize}
    \item $ D= 8 \lambda^{-1} \exp(B +  R) \cdot \frac{\sqrt{n}}{m} \|y-F(0)\|_2 < R = 0.01 \lambda /(n \exp(B)) $
    \item $\| y - F(0)\|_2 = O(\sqrt{n})$
    \item $3n^2\exp(-mR/10)\leq  \delta$
\end{itemize}

Therefore, it suffices to choose:
 \begin{align*}
 m = \Omega( \lambda^{-2} n^2 \exp(4 B) \log(n/\delta) ).
 \end{align*}
\end{proof}

\subsection{Bounding the first order term}\label{sec:induction_for_loss:C_1}
In this section, we bound the first order term $C_1$.
\begin{claim}\label{cla:C1}
If the following conditions hold
\begin{itemize}
    \item Let $B$ be defined as Definition~\ref{def:B}
    \item $C_1 = -2 m\eta (F(t)-y)^\top H(t) (F(t)-y)$
    \item $R\leq 0.01 \lambda / ( n \exp( B) )$
    \item $m = \Omega( \lambda^{-2} \cdot n^2 \cdot \exp(2B) \cdot \sqrt{\log(n/\delta)} )$
\end{itemize}
Then, we have
\begin{align*}
C_1 \leq - m\eta \lambda\cdot \| y - F(t) \|_2^2 
\end{align*}
holds with probability at least $1-(n^2 \cdot \exp(-m R/10)+\delta)$.
\end{claim}

\begin{proof}
By Lemma \ref{lem:perturb_w}, with probability $1-(n^2 \cdot \exp(-m R/10)+\delta)$, we have
\begin{align}\label{eq:upper_lambda}
   & ~ \|H(0)-H(t)\|_F \notag \\
   \leq & ~ 3n R  e^{B}\notag\\
    \leq & ~ \lambda / 4
\end{align}
where the last step follows from choice of $R$ (see Claim Statement).

Given that $\lambda=\lambda_{\min}(H(0))$, by Lemma~\ref{lem:3.1}, we have
\begin{align*}
& ~ \lambda_{\min}(H(t)) \\
\geq & ~ \lambda_{\min}(H(0))- \|H(0)-H(t)\| \\
\geq & ~ \lambda /2.
\end{align*}
where the second step follows from $\lambda_{\min}(H(0))\geq \lambda/2$ and Eq.\eqref{eq:upper_lambda}.

And now we can conclude that
\begin{align*}
  (F(t)-y)^\top H(t) (F(t)-y ) \geq \lambda / 2\cdot\|F(t)-y \|_2^2 .
\end{align*}
\end{proof}

\subsection{Bounding the second order term}\label{sec:induction_for_loss:C_2}
In this section, we bound the second order term $C_2$.

\begin{claim}\label{cla:C2}
If the following conditions hold
\begin{itemize}
    \item $C_2 =  2 \langle y - F(t),v_2\rangle$.
    \item $ R < B $
\end{itemize}
Then we can conclude that
\begin{align*}
C_2 \leq 2 m\eta^2 n \exp(4B) \cdot \| F(t) - y \|_2^2  .
\end{align*}
with probability at least $1-n\cdot \exp(-mR)$.
\end{claim}
\begin{proof}
It holds that
\begin{align*}
C_2 \leq & ~ 2m\eta^2 (F(t)-y)^\top H_{\asy} (F(t)-y) \\
\leq & ~ 2m\eta^2 \| F(t)-y\|_2^2 \cdot \| H_{\asy} \| \\
\leq & ~ 2m\eta^2 \| F(t)-y\|_2^2 \cdot \| H_{\asy} \|_F \\
\leq & ~ 2m\eta^2 \| F(t)-y\|_2^2 \cdot n \| H_{\asy} \|_{\infty} \\
\leq & ~ 2m\eta^2 \| F(t)-y\|_2^2 \cdot n \cdot \exp(4 B )
\end{align*}
where the first step is from $P \in [-1,1]^{m\times n} $, the second step is from Fact~\ref{fac:norm}, the third step is from Fact~\ref{fac:norm}, the forth step follows from Fact~\ref{fac:norm}, the fifth step follows from Claim~\ref{cla:upper_bound_HP}.

\end{proof}

\subsection{Bounding \texorpdfstring{$\| F(t+1) - F(t) \|_2^2$}{}}\label{sec:induction_for_loss:C_3}
In this section, we bound the third order term $C_3$.

\begin{claim}\label{cla:C3}
If the following conditions hold
\begin{itemize}
 \item $C_3  = \| F(t+1) - F(t) \|_2^2$.
 \item $\eta \| \Delta w_r(t) \|_2 \leq 0.01$
 \item $R \leq B$
\end{itemize}
Then with probability at least $1-\delta$, we have
\begin{align*}
C_3 \leq m^2 \eta^2 \cdot n^2 \cdot \exp(8 B) \cdot \| F(t)-y \|_2^2.
\end{align*}
\end{claim}
 
\begin{proof}
According to definition of $F_i(t)$, we have
\begin{align*}
& ~ F_i(t+1) - F_i(t) \\
= & ~   \sum_{r\in [m]} a_r \cdot ( \exp( \langle x_i,w_r(t+1)\rangle ) - \exp(\langle w_r(t),x_i\rangle ) )  \\
= & ~  \sum_{r\in [m]} a_r \cdot \exp(\langle x_i,w_r(t)\rangle) \cdot ( \exp(- \eta 
\langle \Delta  w_r(t),x_i\rangle) - 1 ) 
\end{align*}

Then we have
\begin{align}\label{eq:bound_u_i_k}
|F_i(t+1) - F_i(t) |
\leq & ~ \sum_{r=1}^m \exp(w_r(t)^\top x_i) \cdot | \exp(- \eta \Delta w_r(t)^\top x_i) - 1 | \notag \\
\leq & ~ \sum_{r=1}^m \exp(B+R) \cdot | \exp(- \eta \Delta w_r(t)^\top x_i) - 1 | \notag \\
\leq & ~ \sum_{r=1}^m \exp(B+R) \cdot 2 \eta \| \Delta w_r(t) \|_2 \notag \\
\leq & ~ 2 \eta \exp(B+R) \sum_{r=1}^m \| \Delta w_r(t) \|_2 \notag \\
\leq & ~ 2 \eta \exp(B+R) \sum_{r=1}^m \exp(B + R)  \sqrt{n}\| y - F(t) \|_2 \notag \\
= & ~ 2 m \eta \exp( 2 (B+R) )  \sqrt{n}\| y - F(t) \|_2 
\end{align}     
where the second step is from Lemma~\ref{lem:bound_on_exp_w_and_perturb_w}, the third step is from $\eta \| \Delta w_r(t) \|_2 \leq 0.1 $ and Fact~\ref{fac:taylor}, the fifth step is due to Lemma~\ref{lem:bound_Delta_w_at_time_s}.

We can conclude
\begin{align*}
\| F(t+1) - F(t) \|_2^2 \leq & ~ n \cdot ( 2 m \eta \cdot \exp( 2 (B+R) )  \sqrt{n}\|  F(t)-y \|_2 )^2 \\
\leq & ~ 4 m^2 \eta^2 \cdot n^2 \cdot \exp(4 (B+R) ) \cdot \| F(t)-y \|_2^2 \\
\leq & ~ 4 m^2 \eta^2 \cdot n^2 \cdot \exp( 8 B ) \cdot \| F(t) -y\|_2^2
\end{align*}
where the first step is due to Eq.~\eqref{eq:bound_u_i_k}.
\end{proof}

\ifdefined\isarxiv
%\section*{Acknowledgments}
\bibliographystyle{alpha}
\bibliography{ref}
\else
\bibliography{ref}
\bibliographystyle{alpha}

\fi

\newpage
\onecolumn
\appendix

%%%% Cut-line between first 10 pages and appendix

%%% some writing rules

%% Writing rule for creating tags.
%% Tags :
%% Theorem    \ref{thm:bla_bla}
%% Lemma      \ref{lem:bla_bla}
%% Claim      \ref{cla:bla_bla}
%% Corollary  \ref{cor:bla_bla}
%% Fact       \ref{fac:bla_bla}
%% Definition \ref{def:bla_bla}
%% Section    \ref{sec:bla_bla}
%% Subsection \ref{sub:bla_bla}
%% Equation   \ref{eq:bla_bla}

\end{document}